\documentclass[journal]{IEEEtran}
\usepackage{ifpdf}
\usepackage{cite}
\ifCLASSINFOpdf
   \usepackage[pdftex]{graphicx}
\else
   \usepackage[dvips]{graphicx}
\fi
\usepackage[dvipsnames]{xcolor}
\usepackage{amsmath}
\usepackage{amssymb}
\usepackage{amsthm}
\usepackage{bm}
\usepackage{physics}
\usepackage{dsfont}
\usepackage{nomencl}
\usepackage{etoolbox}
\usepackage[utf8]{inputenc}
\usepackage{multirow}
\usepackage{booktabs}
\usepackage{pbox}
\makeatletter
\def\hlinewd#1{
  \noalign{\ifnum0=`}\fi\hrule \@height #1 \futurelet
   \reserved@a\@xhline}
\makeatother
\usepackage{algorithm}
\usepackage{algorithmic}

\usepackage{array}
\ifCLASSOPTIONcompsoc
 \usepackage[caption=false,font=normalsize,labelfont=sf,textfont=sf]{subfig}
\else
 \usepackage[caption=false,font=footnotesize]{subfig}
\fi
\usepackage{float}
\usepackage{stfloats}
\usepackage{url}
\begin{document}
%
\title{Machine Learning-Driven Virtual Bidding with Electricity Market Efficiency Analysis}
%
%
%
\author{Yinglun~Li,~\IEEEmembership{Student~Member,~IEEE,}
        Nanpeng~Yu,~\IEEEmembership{Senior~Member,~IEEE,}
        and~Wei~Wang,~\IEEEmembership{Member,~IEEE}
\thanks{Y. Li, N. Yu and W. Wang are with the Department of Electrical and Computer Engineering, University of California, Riverside, CA, 92521. Email: nyu@ece.ucr.edu.}
}
\maketitle

\begin{abstract}
This paper develops a machine learning-driven portfolio optimization framework for virtual bidding in electricity markets considering both risk constraint and price sensitivity. The algorithmic trading strategy is developed from the perspective of a proprietary trading firm to maximize profit. A recurrent neural network-based Locational Marginal Price (LMP) spread forecast model is developed by leveraging the inter-hour dependencies of the market clearing algorithm. The LMP spread sensitivity with respect to net virtual bids is modeled as a monotonic function with the proposed constrained gradient boosting tree. We leverage the proposed algorithmic virtual bid trading strategy to evaluate both the profitability of the virtual bid portfolio and the efficiency of U.S. wholesale electricity markets. The comprehensive empirical analysis on PJM, ISO-NE, and CAISO indicates that the proposed virtual bid portfolio optimization strategy considering the price sensitivity explicitly outperforms the one that neglects the price sensitivity. The Sharpe ratio of virtual bid portfolios for all three electricity markets are much higher than that of the S\&P 500 index. It was also shown that the efficiency of CAISO's two-settlement system is lower than that of PJM and ISO-NE.


\end{abstract}
\begin{IEEEkeywords}
Electricity markets, machine learning, virtual bidding, market efficiency.
\end{IEEEkeywords}
%
\IEEEpeerreviewmaketitle
\setlength{\nomitemsep}{-\parsep}
\makenomenclature
\renewcommand\nomgroup[1]{%
  \item[\bfseries
  \ifstrequal{#1}{A}{Variables}{%
  \ifstrequal{#1}{B}{Constants}{}}%
]}
\newcommand{\nomunit}[1]{%
\renewcommand{\nomentryend}{\hspace*{\fill}#1}}
\nomenclature[A, 01]{$\lambda_{i,h}^{DA},\lambda_{i,h}^{RT}$}{DA LMP and RT LMP for node $i$ at hour $h$\quad($\$/MWh$)}
\nomenclature[A, 02]{$\lambda_{i,h}^{dif}$}{Price spread for node $i$ at hour $h$\quad($\$/MWh$)}
\nomenclature[A, 03]{$\bm{\lambda_{h}^{dif}}$}{Vector of price spreads at hour $h$\quad($\$/MWh$)}
\nomenclature[A, 04]{$\lambda_{i,h}^{bid,I},\lambda_{i,h}^{bid,D}$}{Bid prices of INC and DEC for node $i$ at hour $h$\quad($\$/MWh$)}
\nomenclature[A, 05]{$r^I_{i,h},r^D_{i,h}$}{Net profits of INC and DEC for node $i$ at hour $h$\quad($\$$)}
\nomenclature[A, 06]{$z_{i,h}^I,z_{i,h}^D$}{Decision variables of INC and DEC for node $i$ at hour $h$\quad($MWh$)}
\nomenclature[A, 07]{$\bm{z_{h}}$}{Vector of decision variables at hour $h$\quad($MWh$)}
\nomenclature[A, 08]{$u_{i,h}^I,u_{i,h}^D$}{Aggregated virtual bids of INC and DEC from the rest of the market for node $i$ at hour $h$\quad($MWh$)}
\nomenclature[A, 09]{$\bm{u_{h}}$}{Vector of aggregated virtual bids from the rest of the market at hour $h$\quad($MWh$)}
\nomenclature[A, 10]{$x_h$}{Difference between the energy trading company's INC and DEC bidding quantities at hour $h$\quad($MWh$)}
\nomenclature[A, 11]{$y_h$}{Difference between the other market participants' INC and DEC bidding quantities at hour $h$\quad($MWh$)}
\nomenclature[A, 12]{$\lambda_{ref,h}^{dif}$}{Price spread on the system reference node at hour $h$\quad($\$/MWh$)}
\nomenclature[A, 13]{$d_{j,h}$}{Binary variables indicating whether $x_h$ belongs to the $j$-th interval at hour $h$}
\nomenclature[A, 14]{$v_{j,h},w_{j,h},\alpha_h,q_h^k$}{Slack variables introduced in convex relaxation\quad(\$)}
\nomenclature[B, 01]{$\gamma^I,\gamma^D$}{Bidding costs of INC and DEC\quad($\$/MWh$)}
\nomenclature[B, 02]{$prox_{i,h}^I,prox_{i,h}^D$}{Collaterals required by market operators for placing INC and DEC bids for node $i$ at hour $h$\quad($\$/MWh$)}
\nomenclature[B, 03]{$\mathcal{B}$}{Portfolio budget limit\quad(\$)}
\nomenclature[B, 04]{$\mathcal{C}$}{Portfolio risk limit\quad(\$)}
\nomenclature[B, 05]{$a_{j,h}$}{Slope of the linear function defined on the $j$-th interval at hour $h$\quad($\$/MWh^2$)}
\nomenclature[B, 06]{$b_{j,h}$}{Intercept of the linear function defined on the $j$-th interval at hour $h$\quad($\$/MWh$)}
\nomenclature[B, 07]{$c_{j,h}$}{Starting point of the $j$-th interval at hour $h$\quad($MWh$)}
\printnomenclature

\section{Introduction}
%
%
%
%

\IEEEPARstart{T}{he} wholesale electricity markets in the United States operate under the two-settlement system, which comprises of the day-ahead (DA) market and the real-time (RT) market. The DA market clears bid-in supply against bid-in demand and determines DA physical schedules for generators, virtual awards, and DA locational marginal prices (LMPs), which are defined as the marginal costs of serving the next increment of demand at pricing nodes consistent with the existing transmission constraints and performance characteristics of generation resources. The RT market procures ``balancing'' energy to meet the forecast RT grid energy demand and determines RT dispatch signals for resources and RT LMPs.

Electricity price forecasting is one of the most fundamental inputs to decision making problems for  electric utilities and energy trading companies. A general review of the complexity of different electricity price forecasting models with an emphasis on strengths and weaknesses is provided in \cite{WERON20141030}. Widely-used DA and RT LMP forecasting models include auto-regressive integrated moving average (ARIMA) model and its variants \cite{ARIMA2003, forecast1010003}, Markov regime-switching (MRS) model and its variants \cite{WERON200439, windcurtail2013}, and the deep neural network-based models \cite{price2018}.

Electricity markets in the U.S. have two types of bids: physical bids and virtual bids. Physical bids must be backed by physical generation assets, loads, or imports/exports. Virtual bids are financial positions that are not backed by physical assets and do not deliver or consume physical energy. There are two types of virtual bids: increment (INC) offers and decrement (DEC) bids, also known as virtual supply offers and virtual demand bids respectively. INC (DEC) bids sell (buy) energy in the DA market and buy (sell) the same amount of energy back in the RT market.

Virtual bids are introduced in the U.S. electricity markets to drive the price convergence between DA and RT LMPs \cite{testing, mather2017virtual, tang2016model, woo2015virtual}, hedge financial risks \cite{hogan2016virtual}, and increase the market liquidity. The impact of virtual bidding on electricity market has been a controversial topic \cite{hogan2016virtual}. In theory, the introduction of virtual bidding increases market efficiency and reduces price spreads between DA LMP and RT LMP \cite{interconnection2015virtual, Li2015, testing, kazempour2017value}. However, market manipulation \cite{limits, prete2018virtual} and inappropriate market designs such as modeling discrepancies \cite{interconnection2015virtual} and virtual bidding on the interties \cite{arbitrage} can lead to inefficient market solutions.


Researchers tested the hypothesis that the electricity market is efficient by showing if one can find a virtual bid trading strategy that consistently achieves returns in excess of average market returns on a risk-adjusted basis. By identifying profitable virtual bid trading strategy based solely on historical prices, it has been shown that California Independent System Operator (CAISO) \cite{Li2015, testing}, Pennsylvania-New Jersey-Maryland Interconnection (PJM), and New York Independent System Operator (NYISO) \cite{Tong2019} markets fail the weak form of the market efficiency hypothesis test. Most of the prior work \cite{Li2015, testing, Tong2019} underestimate the potential profitability of virtual bid trading strategies and overestimate the market efficiencies by limiting the available information for virtual traders to historical LMPs. Furthermore, they either neglected virtual bid transaction fees \cite{Tong2019} or left out the uplift costs of virtual bids \cite{Li2015} which resulted in overestimation of profitability of virtual bid trading strategies. Our recent work addressed these two issues by developing a machine learning-based algorithmic trading strategy for virtual bidding \cite{wang2019machine}, which uses publicly available information such as load forecasts, meteorological variables, renewable generation forecasts, fuel prices, and historical LMPs to forecast the price spreads between DA and RT LMPs. After accounting for virtual bid transaction fees, uplift costs, and publicly available information, it showed that the algorithmic trading strategy consistently achieved significant profits in the Independent System Operator New England (ISO-NE) market. In other words, the ISO-NE market fails the semi-strong form of market efficiency hypothesis test.


This paper extends the prior works in three ways. First, none of the existing work explicitly models the impacts of virtual bidders' trading activities on electricity market prices \cite{Li2015, testing, Tong2019, wang2019machine}. We hereby develop a machine learning-based estimation algorithm for the LMP spread sensitivity with respect to virtual bid trading quantities. This enables us to develop an algorithmic virtual bidding portfolio optimization framework considering the price sensitivity. By explicitly modeling the impacts of virtual bidding activities on LMPs, the profitability of virtual bid portfolios and market efficiencies can finally be evaluated with different virtual bid market shares. Second, recognizing that the inter-hour operational constraints such as resource ramping constraints have a great influence on LMPs, we accommodate the inter-hour dependencies by adopting a recurrent neural network framework to further improve the existing feedforward neural network-based LMP spread forecasting model \cite{wang2019machine}. Third, most of the prior works perform market efficiency analysis for one electricity market at a time \cite{Li2015, testing, wang2019machine}. This paper performs a large-scale empirical market efficiency analysis across multiple wholesale electricity markets (PJM, ISO-NE, and CAISO).


The unique contributions of this paper are as follows:

$\bullet$ We develop a constrained gradient boosting tree-based algorithm to model the monotonic function representing the LMP spread sensitivity with respect to net virtual bids.

$\bullet$ A virtual bid portfolio optimization framework considering both risk constraints and price sensitivities is established, which is shown to be much more profitable than the version without price sensitivity modeling.


$\bullet$ We develop a neural network-based virtual bid trading quantity forecasting model to predict the hourly difference between market-wide cleared quantities of INC and DEC bids.

$\bullet$ A large-scale empirical market efficiency analysis is conducted for multiple U.S. wholesale electricity markets with respect to different market shares of virtual bids.

The remainder of this paper is organized as follows. Section II formulates the virtual bid portfolio optimization problem with price sensitivities. Section III presents the machine learning-based forecasts for LMP spreads, virtual trading quantities, and the price sensitivity. The empirical study on three U.S. wholesale electricity markets is conducted in Section IV. Section V concludes the paper.

\section{Virtual Bid Portfolio Optimization Problem with Price Sensitivity}
In this section, we formulate the virtual bid portfolio optimization problem with price sensitivity and risk constraint. Note that the virtual trader under consideration is not treated as a price taker. The net profits of virtual bids are modeled in Subsection II.A. The virtual bid portfolio optimization with budget and risk constraints are presented in Subsection II.B. The sensitivity of LMP spread with respect to virtual bid trading quantities is modeled in Subsection II.C. The portfolio optimization problem is reformulated and summarized in Subsections II.D and II.E.

A proprietary trading company engages in virtual bidding activities in wholesale electricity markets through the following process. On a daily basis, the proprietary trading company needs to ensure that it has posted a sufficient amount of collateral in a bank account monitored by the market operator to cover its virtual bid positions. One day before the operating day, the proprietary trading company submits INC offers and DEC bids through the DA market. Then the market operator clears the DA market and returns the virtual bid awards and LMP results back to the proprietary trading company. On the operating day, the virtual bids' positions are automatically liquidated by the market operator in the real-time market, which does not involve any further decision making from the proprietary trading company.


Three modeling assumptions are made in this section. First, it is assumed that the INC offers and DEC bids are guaranteed to be cleared in the DA market, which can be achieved by setting the offer (bid) price to be the price floor (price cap) for INCs and (DECs). Second, we assume that the impact of the virtual bids on congestion patterns in the market is negligible. We ensure that this assumption holds by setting the maximum bid quantity of virtual bids at each node to be 1 MWh. Third, we assume that the spatial-temporal distribution of the LMP spreads does not change much over time.

\subsection{Modeling the Net Profits of Virtual Bids}
Let $\lambda_{i,h}^{DA}$ and $\lambda_{i,h}^{RT}$ denote the DA and RT LMP for node $i$ at hour $h$. The price spread $ \lambda_{i,h}^{dif} $ for node $i$ at hour $h$ is defined as the difference between DA and RT LMP, $\lambda_{i,h}^{dif}=\lambda_{i,h}^{DA}-\lambda_{i,h}^{RT} $.
$ \lambda_{i,h}^{bid,I} $ and $ \lambda_{i,h}^{bid,D} $ are bid prices of INC and DEC for node $ i $ at hour $ h $. Note that INCs are cleared when $ \lambda_{i,h}^{bid,I}\leq\lambda_{i,h}^{DA} $ and DECs are cleared when $ \lambda_{i,h}^{bid,I}\geq\lambda_{i,h}^{DA} $. The bidding costs associated with INCs and DECs are denoted as $ \gamma^I $ and $ \gamma^D $. The bidding costs include uplift cost and transaction fee. The expected net profit of the INC offer ($ r_{i,h}^I $) and the DEC bid ($ r_{i,h}^D $) for node $i$ at hour $h$, can be calculated as:
\begin{align}
    E[r_{i,h}^I]&=E[(\lambda_{i,h}^{dif}-\gamma^I)\mathds{1}(\lambda_{i,h}^{bid,I}\leq\lambda_{i,h}^{DA})] \label{eq:net_profit_inc}\\
    E[r_{i,h}^D]&=E[(-\lambda_{i,h}^{dif}-\gamma^D)\mathds{1}(\lambda_{i,h}^{bid,D}\geq\lambda_{i,h}^{DA})] \label{eq:net_profit_dec}
\end{align}

In this paper, it is assumed that INCs and DECs are submitted into the DA market with bidding prices that will guarantee their clearance. This can be achieved by setting the bid price to be the price floor (price cap) for INCs (DECs).


\subsection{Virtual Bid Portfolio Optimization with Budget and Risk Constraints Considering Price Sensitivity}

The objective of an energy trading company is to develop a portfolio of virtual bids, which maximizes its profit with limited risks. The LMPs will be impacted by the submitted virtual bids. Thus, the price sensitivity with respect to the virtual bid trading quantities should be considered in the portfolio optimization process. In some related work like \cite{Attarha2019}, the trader is assumed to be a price-taker in order to comply with the assumption that its bids are guaranteed to be cleared. In fact, to ensure the virtual bids are cleared, the trader can either make its bids self-schedules or setting bid price to the price floor or the price cap for supply offers or demand bids respectively. The price sensitivity assumption is more realistic and does not necessarily affect the bid clearance.

The trading quantities of INC ($ z^I_{i,h} $) and DEC ($ z^D_{i,h} $) for node $i$ at hour $h$ across the operating day are the decision variables. To alleviate the impact of virtual bids on the congestion patterns in the electricity market, we assume that $ z^I_{i,h} $ or $ z^D_{i,h} $ are binary variables, where 0 represents no virtual bids and 1 represents a 1 MWh of virtual bid.

The portfolio optimization problem of virtual bidding is formulated as follows:
\begin{equation}
    max_{\bm{z}}\;\sum^{24}_{h=1}\sum^N_{i=1}(z^I_{i,h}E[r^I_{i,h}(\bm{z_h})]+z^D_{i,h}E[r^D_{i,h}(\bm{z_h})]) \label{formula:opt_objective}
\end{equation}
\begin{gather}
 s.t. \quad \sum^N_{i=1}\sum^{24}_{h=1}(z^I_{i,h}prox^I_{i,h}+z^D_{i,h}prox^D_{i,h})\leq\mathcal{B} \label{ineq:bugdet}\\
    \sum^{24}_{h=1}CVaR_{\beta}(f_h(\bm{z_h},\bm{\lambda^{dif}_h}))\leq\mathcal{C} \label{ineq:risk}\\
    f_h(\bm{z_h},\bm{\lambda^{dif}_h})=-\sum^N_{i=1}(z^I_{i,h}r^I_{i,h}+z^D_{i,h}r^D_{i,h}) \label{eq:financial_loss}
\end{gather}
where $ prox^I_{i,h} $ and $ prox^D_{i,h} $ are the collaterals required by market operators for placing INC and DEC bids. $ \mathcal{B} $ is the portfolio budget limit. $ \mathcal{C} $ is the portfolio risk limit. The conditional value-at-risk (CVaR) is used to quantify the financial risk of the virtual bid portfolio as in (\ref{ineq:risk}), where $ \beta $, $ \bm{z_h} $, $\bm{\lambda^{dif}_h}$, and $f_h(\bm{z_h},\bm{\lambda^{dif}_h}) $ are the confidence level associated with CVaR, the vector of decision variables, the vector of LMP spreads , and the portfolio loss at hour $h$ respectively. 



By explicitly specifying the impacts of virtual bids submitted by the energy trading company on LMP spreads, we can rewrite $E[r^I_{i,h}(\bm{z_h})]$ and $E[r^D_{i,h}(\bm{z_h})]$ as
\begin{align}
    E[r_{i,h}^I(\bm{z_h})]&=E[(\lambda_{i,h}^{dif}(\bm{z_h, u_h})-\gamma^I)\mathds{1}(\lambda_{i,h}^{bid,I}\leq\lambda_{i,h}^{DA})] \label{eq:net_profit_inc_sensitivity}\\
    E[r_{i,h}^D(\bm{z_h})]&=E[(-\lambda_{i,h}^{dif}(\bm{z_h, u_h})-\gamma^D)\mathds{1}(\lambda_{i,h}^{bid,D}\geq\lambda_{i,h}^{DA})] \label{eq:net_profit_dec_sensitivity}
\end{align}
where $\bm{u_h}$ denotes the vector of aggregated virtual bids from the rest of the electricity market at hour $h$. In other words, the LMP spread is influenced by the virtual bidding activities of the energy trading company under consideration and the rest of the market participants.
\subsection{Sensitivity of LMP Spreads with Respect to Virtual Bid Trading Quantities}
The impacts of the virtual bidding activities from the energy trading company and the rest of the market participants on LMP spreads can be approximated as: 
\begin{equation}
    \lambda_{i,h}^{dif}(\bm{z_h, u_h}) = \lambda_{i,h}^{dif}(x_h+y_h) \label{eq:lmp_diff_bid_quant}
\end{equation}

The difference between the energy trading company's INC and DEC trading quantities at hour $h$  ($ x_h $) and that of the other market participants' ($y_h$) are defined as:
\begin{align}
    x_h=\sum^N_{i=1}z^I_{i,h}-\sum^N_{i=1}z^D_{i,h} \label{eq:bid_quant_trader}\\
    y_h=\sum^N_{i=1}u^I_{i,h}-\sum^N_{i=1}u^D_{i,h} \label{eq:bid_quant_market}
\end{align}
where $u^I_{i,h}$ and $u^D_{i,h}$ are the aggregated INC and DEC bids of the rest of the market for node $i$ at hour $h$ respectively.

It is extremely difficult to estimate the impact of virtual bidding on individual node's price spread due to the lack of nodal virtual bid trading quantity. Thus, the impacts of virtual bidding on an individual node is approximated by the impacts on the system reference node:
\begin{equation}
    \lambda_{i,h}^{dif}(x_h+y_h) \approx \lambda_{i,h}^{dif}(y_h) + [ \lambda_{ref,h}^{dif}(x_h+y_h) - \lambda_{ref,h}^{dif}(y_h) ] \label{eq:lmp_diff_approx}
\end{equation}

Note that $y_h$ in equation (\ref{eq:lmp_diff_approx}) is determined outside the energy trading companies' portfolio optimization problem and can be estimated with a machine learning model.

To make the portfolio optimization problem tractable, we model the impact of the energy trading company's virtual bids on the market reference LMP as a piece-wise linear function of $x_h$ shown in Figure 1. 
\begin{gather}
    \lambda_{ref,h}^{dif}(x_h+y_h) - \lambda_{ref,h}^{dif}(y_h)=\sum^{M_h}_{j=1}(a_{j,h}x_h+b_{j,h})d_{j,h} \label{eq:lmp_ref_function}\\
    c_{j,h}-S(1-d_{j,h})\leq x_h\leq c_{j+1,h}+S(1-d_{j,h}) \label{ineq:interval_range}\\
    \sum^{M_h}_{j=1}d_{j,h}=1 \label{eq:interval_indicator}\\
    \underline{x_h}\leq x_h\leq\overline{x_h} \label{ineq:x_valid_range}
\end{gather}
\begin{figure}[!t]
    \centering
    \includegraphics[width=3.5in]{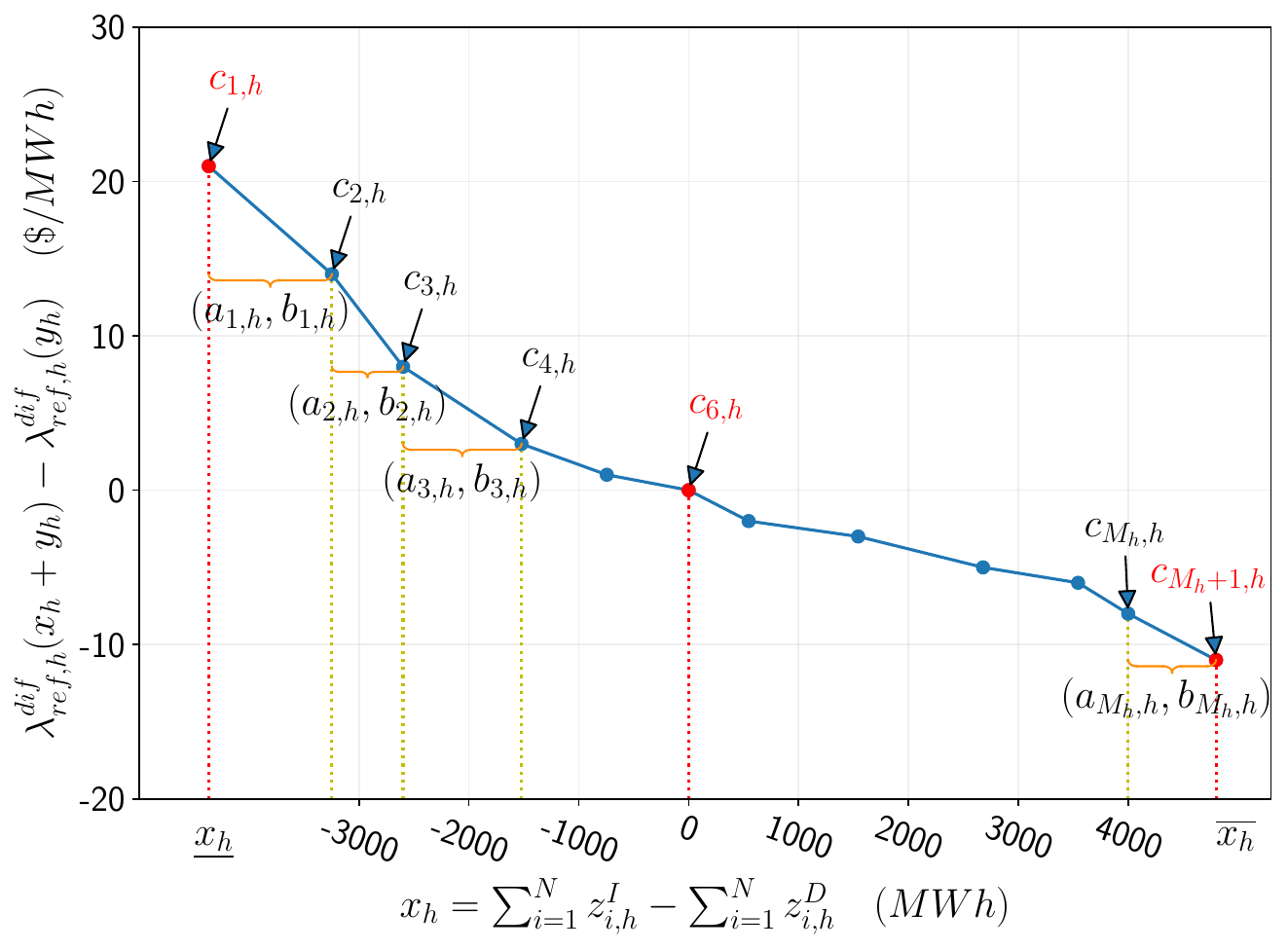}
    \caption{Illustration of the piece-wise linear LMP sensitivity function.}
    \label{fig0}
\end{figure}
Here $ \underline{x_h} $ and $ \overline{x_h} $ are the lower and upper bound of $ x_h $, which are selected based on historical minimum and maximum hourly market-wide trading quantity of INCs minus DECs. Starting from the lower bound, we number the intervals in ascending order from $j=1$ to $M_h$ which hits the upper bound. $c_{j,h}$ is the impact of virtual bids on market reference LMP at the starting point of $ j $-th interval. $S$ is a sufficiently large real number. Let $ a_{j,h} $ and $ b_{j,h} $ denote the slope and intercept of the linear function defined on $j$-th interval. The binary variable $ d_{j,h} $ indicates whether $ x_h $ belongs to $j$-th interval at hour $h$. Equation (\ref{ineq:interval_range}) enforces that $ x_h $ can only fall in a single interval. The parameters of the piece-wise linear function are derived from a gradient boosting tree (GBT) model, which will be discussed in detail in Section III.C.

Note that $a_{j,h}$ should be less than 0 for all intervals. This is because as the INC (DEC) trading quantity of the energy trading company increases, the hourly LMP spread at the reference node decreases (increases). We will describe how to enforce the piece-wise linear function to be monotonically decreasing by using a GBT model in Section III.C. Finally, it should be noted that parameters of the piecewise linear function depend on $y_h$, the difference between INC and DEC trading quantities of the other market participants at hour $h$ and other factors that affect market clearing outcomes.

\subsection{Reformulation of Portfolio Optimization Problem}
By substituting equations (\ref{eq:net_profit_inc_sensitivity}), (\ref{eq:net_profit_dec_sensitivity}), (\ref{eq:lmp_diff_bid_quant}), (\ref{eq:lmp_diff_approx}), and (\ref{eq:lmp_ref_function}) into (\ref{formula:opt_objective}), the objective function of the portfolio optimization problem can be rewritten as:
\begin{multline}
    \sum^{24}_{h=1} \big\{ \sum^N_{i=1}z^I_{i,h}E[(\lambda_{i,h}^{dif}-\gamma^I)]+\sum^N_{i=1}z^D_{i,h}E[(-\lambda_{i,h}^{dif}-\gamma^D)]\\
    +E[\sum^{M_h}_{j=1}(a_{j,h}x^2_h+b_{j,h}x_h)d_{j,h}] \big\} \label{formula:opt_objective_sensitivity}
\end{multline}

To make the objective function concave, we introduce slack variables $v_{j,h}$ and rewrite the equation (17) as:
\begin{gather}
    \sum^{24}_{h=1} \big\{ \sum^N_{i=1}z^I_{i,h}E[(\lambda_{i,h}^{dif}-\gamma^I)] \nonumber \\ 
    +\sum^N_{i=1}z^D_{i,h}E[(-\lambda_{i,h}^{dif}-\gamma^D)]+E[\sum^{M_h}_{j=1}v_{j,h}d_{j,h}] \big\} \label{formula:opt_objective_sensitivity_relax_1} \\ 
    s.t. \quad v_{j,h}\leq a_{j,h}x^2_h+b_{j,h}x_h \; \forall j,h \label{ineq:relax_1}
\end{gather}

Note that the term $ E[\sum^{M_h}_{j=1}v_{j,h}d_{j,h}] $ still makes the objective function non-concave. The objective function is further relaxed by introducing additional slack variables $w_{j,h}$ as:
\begin{multline}
    \sum^{24}_{h=1} \big\{ \sum^N_{i=1}z^I_{i,h}E[(\lambda_{i,h}^{dif}-\gamma^I)]\\
    +\sum^N_{i=1}z^D_{i,h}E[(-\lambda_{i,h}^{dif}-\gamma^D)]+E[\sum^{M_h}_{j=1}w_{j,h}] \big\} \label{formula:opt_objective_sensitivity_relax_2}
\end{multline}
\begin{gather}
    s.t. \quad -Sd_{j,h}\leq w_{j,h}\leq Sd_{j,h} \; \forall j,h \label{ineq:relax_2}\\
    -S(1-d_{j,h})\leq w_{j,h}-v_{j,h}\leq S(1-d_{j,h}) \; \forall j,h \label{ineq:relax_3}
\end{gather}
\subsection{Summary of the Portfolio Optimization Formulation}
In summary, the virtual bid portfolio optimization problem can be formulated as follows:
\begin{gather*}
    max_{\bm{z}}\;(\ref{formula:opt_objective_sensitivity_relax_2})\\
    s.t. \quad (\ref{ineq:bugdet})-(\ref{eq:financial_loss}), (\ref{eq:bid_quant_trader}), (\ref{eq:bid_quant_market}), (\ref{ineq:interval_range})-(\ref{ineq:x_valid_range}), (\ref{ineq:relax_1}), (\ref{ineq:relax_2}), (\ref{ineq:relax_3})
\end{gather*}
This is a mixed-integer quadratically-constrained programming problem and can be solved by optimization engines such as CPLEX\cite{Bliek2014SolvingMQ}.

\subsection{Formulation of Risk Constrained Portfolio Optimization}
In this work, we adopt CVaR as the risk measure for the virtual bid portfolio. It is chosen as the preferred risk measure because it is not only a coherent measurement of risk, but also accurately captures the tail distribution of portfolio loss function. In order to introduce CVaR, we need to first define value-at-risk (VaR)\cite{RePEc:bla:mathfi:v:9:y:1999:i:3:p:203-228}. Let us first define the probability of the loss $ f_h(\bm{z_h},\bm{\lambda^{dif}_h}) $ at hour $ h $ not exceeding $ \alpha_h $ as:
\begin{equation}
    \Psi(\bm{z_h},\alpha_h)=\int_{f_h(\bm{z_h},\bm{\lambda^{dif}_h})\leq\alpha_h}p(\bm{\lambda^{dif}_h})d\bm{\lambda^{dif}_h} \label{eq:CDF_loss}
\end{equation}

Here $ p(\bm{\lambda^{dif}_h}) $ is the density function of LMP spread vector. $ \Psi(\bm{z_h},\alpha_h) $ is the cumulative distribution function of the portfolio loss associated with decision vector $ \bm{z_h} $. The $ VaR_{\beta} $ of the portfolio is the minimum portfolio loss such that the probability of having a smaller loss is $\beta$:
\begin{equation}
    VaR_{\beta}(\bm{z_h})=min\{\alpha_h:\Psi(\bm{z_h},\alpha_h)\geq\beta\} \label{eq:VaR}
\end{equation}

The CVaR of the portfolio $ CVaR_{\beta} $ is defined as the expected loss given that the loss is no less than $ VaR_{\beta} $:
\begin{align}
    &CVaR_{\beta}(f_h(\bm{z_h},\bm{\lambda^{dif}_h}))\nonumber\\
    &=\dfrac{1}{1-\beta}\int_{f_h(\bm{z_h},\bm{\lambda^{dif}_h})\geq VaR_{\beta}(\bm{z_h})}f_h(\bm{z_h},\bm{\lambda^{dif}_h})p(\bm{\lambda^{dif}_h})d\bm{\lambda^{dif}_h} \label{eq:CVaR}
\end{align}

It has been proved that $ CVaR_{\beta} $ is upper bounded by the function $ F_{\beta}(\bm{z_h},\alpha_h) $ \cite{Rockafellar00optimizationof}:
\begin{multline}
    F_{\beta}(\bm{z_h},\alpha_h)=\alpha_h\\
    +\dfrac{1}{1-\beta}\int_{\bm{\lambda^{dif}_h}}[f_h(\bm{z_h},\bm{\lambda^{dif}_h})-\alpha_h]^+p(\bm{\lambda^{dif}_h})d\bm{\lambda^{dif}_h} \label{eq:CVaR_upper}
\end{multline}

In other words, CVaR can be represented as:
\begin{gather}
    CVaR_{\beta}\big(f_h(\bm{z_h},\bm{\lambda^{dif}_h})\big) = min_{\alpha_h}\;F_{\beta}(\bm{z_h},\alpha_h) \label{eq:CVaR_bound}
\end{gather}

Under the assumption that the spatial-temporal distribution of the LMP spread does not change much over time, $F_{\beta}(\bm{z_h},\alpha_h)$ can be approximated by Monte Carlo sampling with historical LMP spread samples as:
\begin{equation}
    F_{\beta}(\bm{z_h},\alpha_h)=\alpha_h+\dfrac{1}{(1-\beta)N_s}\sum_{k=1}^{N_s}[f_h(\bm{z_h},\bm{\lambda^{dif}_{h,k}})-\alpha_h]^+ \label{eq:CVaR_upper_approx}
\end{equation}
where $N_s$ is the number of historical LMP spread samples.

To remove the $ max(0,\,x) $ function on the last term, \eqref{eq:CVaR_upper_approx} can be further relaxed as:
\begin{equation}
    F_{\beta}(\bm{q_h},\alpha_h)=\alpha_h+\dfrac{1}{(1-\beta)N_s}\sum_{k=1}^{N_s}q_h^k \label{eq:CVaR_upper_approx_relax}
\end{equation}
\begin{gather}
    s.t. \quad q_h^k\geq f_h(\bm{z_h},\bm{\lambda^{dif}_{h,k}})-\alpha_h \label{ineq:relax_4}\\
    q_h^k\geq 0 \label{ineq:relax_5}
\end{gather}

Then, the portfolio optimization problem in Section II can be reformulated as:
\begin{gather}
    max_{\bm{z},\bm{q},\bm{\alpha}}\;(\ref{formula:opt_objective_sensitivity_relax_2})\nonumber\\
    s.t. \quad \sum^{24}_{h=1}F_{\beta}(\bm{q_h},\alpha_h)\leq\mathcal{C} \label{ineq:budget_upper}\\
    (\ref{ineq:bugdet}), (\ref{eq:financial_loss}), (\ref{eq:bid_quant_trader}), (\ref{eq:bid_quant_market}), (\ref{ineq:interval_range})-(\ref{ineq:x_valid_range}), (\ref{ineq:relax_1}), (\ref{ineq:relax_2}), (\ref{ineq:relax_3}), (\ref{eq:CVaR_upper_approx_relax})-(\ref{ineq:relax_5})\nonumber
\end{gather}

\section{Data-Driven Forecasting for Virtual Bidding}
\subsection{Neural Network based LMP Spread Forecast}
The LMP spread between DA and RT market $\lambda_{i,h}^{dif}$ is a highly nonlinear function of explanatory variables such as meteorological variables at key weather stations, fuel price forecasts, zonal load forecasts, and renewable generation forecasts. It has been shown that feedforward neural networks such as multilayer perceptrons (MLP) and mixture density networks are quite effective in learning the nonlinear function \cite{wang2019machine}. In feedforward neural networks, the training samples at different hours are considered to be independent. However, the 24 hours of LMPs of the DA market are determined jointly via the security constrained unit commitment (SCUC) and security constrained economic dispatch (SCED) processes in practice. In particular, many inter-hour operational constraints such as resource ramping constraints are enforced in SCUC and SCED. For example, it is more likely to observe spikes in LMPs when the net-load of the hour and the increase in net-load from the previous hour are both very high.

To accommodate the inter-hour dependencies, we decide to adopt the long short-term memory \cite{doi:10.1162/089976600300015015} network, which is capable of learning long-term dependencies in the data. We use the cell state of LSTM to carry electricity market operation status information. LSTM employs three types of gates to control the information flow. The forget gate and input gate control which information should be discarded and added to the cell state. The output gate influences how the information in the cell state is used to predict the LMP spread.

In addition to the typical input feature normalization, we also need to perform special preprocessing for the target variables, i.e., the LMP spreads. Note that the LMP spreads are extremely volatile and have many spikes. If the output LMP spread is not scaled to flatten its distribution, then the LMP spikes will dominate the loss function of the neural network. This essentially makes all other training samples ineffective. To mitigate this problem, we leverage the parameterized sigmoid function $ f(x) = \frac{1}{1+e^{-x/\theta}} $ to scale the LMP spread to $ (0,\,1) $, with a hyper-parameter $\theta$. The activation functions are chosen to be hyperbolic tangent function ``tanh" for the hidden layers and the sigmoid function for the output layer.
\subsection{Neural Network based Virtual Bid Trading Quantity Forecast}
To estimate the impact of the energy trading company's virtual bids on LMP spreads, we need to first forecast the aggregated virtual trading quantities $y_h$ from the rest of the market participants. An MLP is adopted to address this problem. Similar to the LMP spread forecast model, the inputs to the MLP also consist of hourly market-wide features such as zonal load forecast, wind and solar generation forecast, meteorological variables, and one-hot encoding for trading hour. The output of the neural network is the hourly difference between market-wide cleared quantities of INC and DEC bids. These features can be found in the data archives maintained by the market operators. For this regression task, the typical input feature normalization and target scaling are required as well. As discussed in Section III.A, we leverage a parameterized sigmoid function to scale the LMP spreads. Similarly, here we apply the sigmoid scaling function $ f(x) = \frac{1}{1+e^{-x/\theta_v}} $ to scale the target, which is the market-wide cleared virtual trading quantity of INCs minus DECs. The hyper-parameter $\theta_v$ is much greater than $\theta$, which is used in the scaling function of the LMP spread. This is because the range of the virtual bid trading quantity is much wider than that of the LMP spread.



\subsection{Constrained Gradient Boosting Tree based Price Sensitivity Modeling}

We expand the XGBoost \cite{DBLP:journals/corr/ChenG16} method to model the monotonic piece-wise linear function representing the LMP spread sensitivity with respect to the net virtual bids. XGBoost is a variation of the gradient boosting tree method \cite{10.2307/2699986}. It is selected as the base model due to its scalability and capability to handle sparse datasets. The modification we make to the algorithm ensures that the learned LMP spread sensitivity function is monotonic.

As a supervised learning model, XGBoost searches in the space of regression trees to minimize a regularized objective. XGBoost addresses the optimization problem over the function space by additive training. This incremental approach, also called ``tree boosting", helps to learn the tree structure with the optimal score. Specifically, at the $t$-th iteration, a new tree $f_t$ is added to optimize the objective:
\begin{gather*}
    \mathcal{L}^{(t)}=\sum_{i=1}^{N}l(\bm{y_i},\hat{\bm{y_i}}^{(t-1)}+f_t(\bm{x_i}))+\Omega(f_t)
\end{gather*}
where $N$ is the number of training samples, $\bm{x_i}$ is the $i$-th input, $\bm{y_i}$ is the actual output, $\hat{\bm{y_i}}$ is the predicted output, and $\Omega(f_t)$ is the regularization term. On top of the inputs for the neural network based LMP spread forecast model, we add the market-wide net INC offer quantity as an additional input for the XGBoost based model. The outputs are the LMP spreads at the reference node.

By taking Taylor expansion at $(\bm{y_i},\hat{\bm{y_i}}^{(t-1)})$, the second-order approximation of the objective can be derived. For a fixed tree structure $q(\bm{x})$, the optimal leaf weight $w^*_j$ and objective can be derived as:
\begin{gather*}
    \mathcal{L}^{*(t)}=-\dfrac{1}{2}\sum_{j=1}^{T}\dfrac{{G_j}^2}{H_j+\lambda}+\gamma T\\
    w^*_j=-\dfrac{G_j}{H_j+\lambda}
\end{gather*}
where $G_j$ and $H_j$ denote the sum of first-order derivatives $g_i$ and the sum of second-order derivatives $h_i$ of leaf $j$ respectively. $T$ is the total number of leaves in the tree. This score can be regarded as a quality measure of the tree structure at $t$-th iteration. When a new split is made, we can calculate the change of the score, also called the gain, as follows:
\begin{gather*}
    \mathcal{L}_{split}=\dfrac{1}{2}[\dfrac{G_L^2}{H_L+\lambda}+\dfrac{G_R^2}{H_R+\lambda}-\dfrac{(G_L+G_R)^2}{H_L+H_R+\lambda}]-\gamma
\end{gather*}

Note that after a new split is made, the left leaf weight is not guaranteed to be higher than that of the right leaf. To ensure the learned LMP spread function is monotonically decreasing as a function of the net INC quantity, we modify the algorithm for finding split points in the XGBoost framework. The proposed greedy algorithm for finding splits that ensures monotonicity of the learned function is summarized in Algorithm 1.

The inputs to the algorithm include the dimension of the input feature space ($D$), the index of the input feature involved with the monotonicity constraint ($p$), and the combination set ($I$) of indices of input samples assigned to each leaf. The indices assigned to leaf $j$ is defined as $I_j=\{i\mid q(\bm{x_i})=j\}$, where $q(\bm{x_i})$ represents the mapping function that assigns the $i$-th sample to the $j$-th leaf.

Starting from line 3, the algorithm searches through all $D$ features for split candidates. For each feature $k$, the algorithm searches through all possible split points in an ascending order. If the feature is the one that is involved in the monotonicity constraint, then the weight of the left leaf must be higher than that of the right leaf, before the split's gain in objective is saved and recommended as a candidate split. Finally, the candidate split that satisfies the monotonicity constraint with highest gain is selected.

\begin{algorithm}
 \caption{Greedy Algorithm for Finding Splits for Learning Monotonically Decreasing Functions.}
 \begin{algorithmic}[1]
 \REQUIRE $I$, $D$, $p$ 
 \ENSURE  Split with best gain and monotonicity constraint
  \STATE $gain\leftarrow0$
  \STATE $G\leftarrow\sum_{i\in I}g_i$, $H\leftarrow\sum_{i\in I}h_i$
  \FOR {$k = 1$ to $D$}
  \STATE $G_L\leftarrow0$, $H_L\leftarrow0$
  \FOR {$j$ in $Ascending \; sort(I, by\;x_{jk})$}
  \STATE $G_L\leftarrow G_L+g_j$, $H_L\leftarrow H_L+h_j$
  \STATE $G_R\leftarrow G-G_L$, $H_R\leftarrow H-H_L$
  \IF {($k == p$)}
  \IF {($-\frac{G_L}{H_L+\lambda}\geq-\frac{G_R}{H_R+\lambda}$)}
  \STATE $gain\leftarrow$
  \STATE $\qquad max(gain, \frac{G_L^2}{H_L+\lambda}+\frac{G_R^2}{H_R+\lambda}-\frac{(G_L+G_R)^2}{H_L+H_R+\lambda})$
  \ENDIF
  \ELSE
  \STATE $gain\leftarrow max(gain, \frac{G_L^2}{H_L+\lambda}+\frac{G_R^2}{H_R+\lambda}-\frac{(G_L+G_R)^2}{H_L+H_R+\lambda})$
  \ENDIF
  \ENDFOR
  \ENDFOR
 \RETURN Split with the highest feasible gain
 \end{algorithmic}
\end{algorithm}

With a finite number of splits, the output of the modified XGBoost model is a step function. To convert this function into the piece-wise linear function in equation (13), we can simply connect the adjacent splitting points. The number of intervals $M_h$ of the piece-wise linear function in equation (13) is determined by the final output of the modified XGBoost algorithm. 
\section{Numerical Study}
\subsection{Setup for Numerical Study}
We validate the proposed algorithmic trading strategy and perform market efficiency analysis on three wholesale electricity markets in the US: Pennsylvania-New Jersey-Maryland Interconnection (PJM), California Independent System Operator (CAISO) and ISO New England (ISO-NE). Three years of historical data are collected for each market. The first year of data is used for initial model training and the last two years of data is used for rolling forecast. PJM and ISO-NE's historical data ranges from January 2015 to December 2017. CAISO's historical data ranges from July 2018 to June 2020. We perform rolling forecasts and update the forecast model on a monthly basis. The amount of training data is always kept at one year. The common inputs to deep neural networks and gradient boosting trees for all three electricity markets include load forecast, meteorological variables (temperature, humidity, wind speed, and precipitation), fuel price, and one-hot encoding for operating hour. The ISO-NE's models include an additional input: wind generation forecast. The CAISO's models include three extra inputs: estimated import, wind, and solar generation forecast. Furthermore, we conduct the forecasting task using only the common inputs for all three markets to show the impacts on forecasting performance from the additional input features for ISO-NE and CAISO.
The architectural hyperparameters of the MLP and LSTM models are summarized in Table \ref{table:Structure}. For the LSTM model, the first two numbers shown in the table represent the dimensions of two stacked LSTM layers. The other numbers in the list represent the dimensions of the remaining fully-connected layers. The first LSTM layer returns the sequence of all hidden states while the second LSTM layer simply returns a single output at the last time step. Both neural networks employ the hyperbolic tangent function (tanh) as the activation function for the hidden layers. Dropout is introduced to regularize both neural networks. The dropout rate is selected to be 20\%. The learning rate is set at 0.001. The Adam optimizer is used for both neural network models. The training batch sizes are 2048 for both LSTM and MLP models. Finally, the LMP spread scaling parameter $\theta$ in the sigmoid function is set to be 20 for CAISO and 10 - 40 for PJM and ISO-NE depending on the price spread volatility of the node.

In this work, all the scripts are written in \emph{Python 3.7}. The neural network models are based on \emph{Tensorflow 1.15} software platform and \emph{Nvidia RTX 2080 Ti} hardware platform. The portfolio optimization is based on \emph{CVXPY 1.1} and the optimizer used is \emph{Gurobi 9.1}.

\begin{table}[!t]
    \setlength\extrarowheight{2pt}
    \renewcommand{\arraystretch}{1.2}
    \caption{Architecture Hyperparameters of Neural Networks}
    \label{table:Structure}
    \centering
    \begin{tabular}{c|c|c}
    \hlinewd{1pt}
    Model Type & Market & Hidden Units \\
    \hlinewd{1pt}
    \multirow{3}{*}{MLP} & PJM & [128, 64, 32] \\ \cline{2-3}
                         & ISO-NE & [64, 32] \\ \cline{2-3}
                         & CAISO & [128, 64, 32] \\
    \hlinewd{1pt}
    \multirow{3}{*}{LSTM} & PJM & [64, 128, 128, 64, 32] \\ \cline{2-3}
                         & ISO-NE & [32, 64, 64, 32] \\ \cline{2-3}
                         & CAISO & [64, 128, 128, 64, 32] \\
    \hlinewd{1pt}
    \end{tabular}
\end{table}

\subsection{Performance Comparison of LMP Spread Forecasting Algorithms}

The LMP spreads between DA and RT markets quantify the potential net revenue of virtual bids without considering trading costs. Therefore, we forecast the LMP spreads directly instead of forecasting DA and RT LMPs separately. To better compare the performance of different LMP spread forecasting models, we introduce two evaluation metrics, which are tailored for the virtual bidding setup.

The first evaluation metric quantifies if the LMP spreads forecast led the virtual trader to place a virtual bid in the right direction. If the LMP spread forecast correctly forecasted that the RT LMP is higher than DA LMP, then the virtual trader will place a DEC bid that is profitable. The LMP spread forecast accuracy metric is formally defined as the proportion of the time that the LMP spread predictions have the correct sign. To better quantify the capability of the LMP spread forecasting algorithms to capture spikes that lead to massive profit, the forecasting accuracy is evaluated on spikes that are in the top 1th percentile of absolute LMP spreads. The second evaluation metric is the root mean square error (RMSE) of the LMP spread forecasts for the top 1th percentile of absolute LMP spreads.

We compared the LMP spread forecasting performance of the proposed MLP and LSTM models with a benchmark support vector regression machine (SVRM) \cite{Drucker97supportvector}. The performance metrics are calculated by averaging the results on three electricity markets and across 2 years. The LMP spread spike forecast accuracies for SVRM, MLP and LSTM are 40.21\%, 51.78\%, and 56.18\%. The RMSE for the LMP spread spike forecasts for SVRM, MLP and LSTM are \$217.35/MWh, \$216.15/MWh, and \$214.98/MWh. The LSTM-based LMP spread forecasting algorithm outperforms the MLP and SVRM-based approaches.

As mentioned in Subsection IV.A, we also conduct the forecasting task again using the common input features instead of different ones for all three markets using the LSTM models. The removal of additional input features for ISO-NE and CAISO lowers the LMP spread spike forecast accuracy by 3.96\% and increases the RMSE for LMP spread forecast by 0.20\%. This result shows that these additional input features are helpful in improving the LMP spread forecast performance.


\subsection{Profitability of Algorithmic Trading Strategies Without Considering Price Sensitivity}
We first evaluate the profitability of algorithmic trading strategy without considering the impacts of the trading company's virtual bids on LMP spreads. The evaluation is conducted over a 2-year period. The daily virtual bid portfolio budget limits in the 1st-year are set to \$600K, \$25K, and \$85K for PJM, ISO-NE and CAISO, respectively. In the 2nd-year, the daily budget limit for PJM is altered to \$330K, while the other two remain the same. The selected budget limits represent approximately 5\% of the market share for virtual bids in the corresponding markets for that year. The virtual bid market share is defined as the portion of the market-wide cleared virtual trading quantity controlled by the proprietary trading company. The daily portfolio risk limit is set to be the same as the budget limit. As explained in subsection II.C, the upper and lower bound of $x_h$ are selected based on historical minimum and maximum hourly market-wide bidding quantity of INCs minus DECs in the corresponding wholesale electricity markets. Specifically, for PJM, ISO-NE, and CAISO, the upper and lower bounds $(\underline{x_h}, \overline{x_h})$ in MWh are (-7812, 6821), (-331, 378), and (-3525, 3450) for the first testing year and (-8815, 5622), (-255, 435), and (-5647, 2798) for the second testing year.

\begin{figure}[!t]
    \centering
    \includegraphics[width=3.5in]{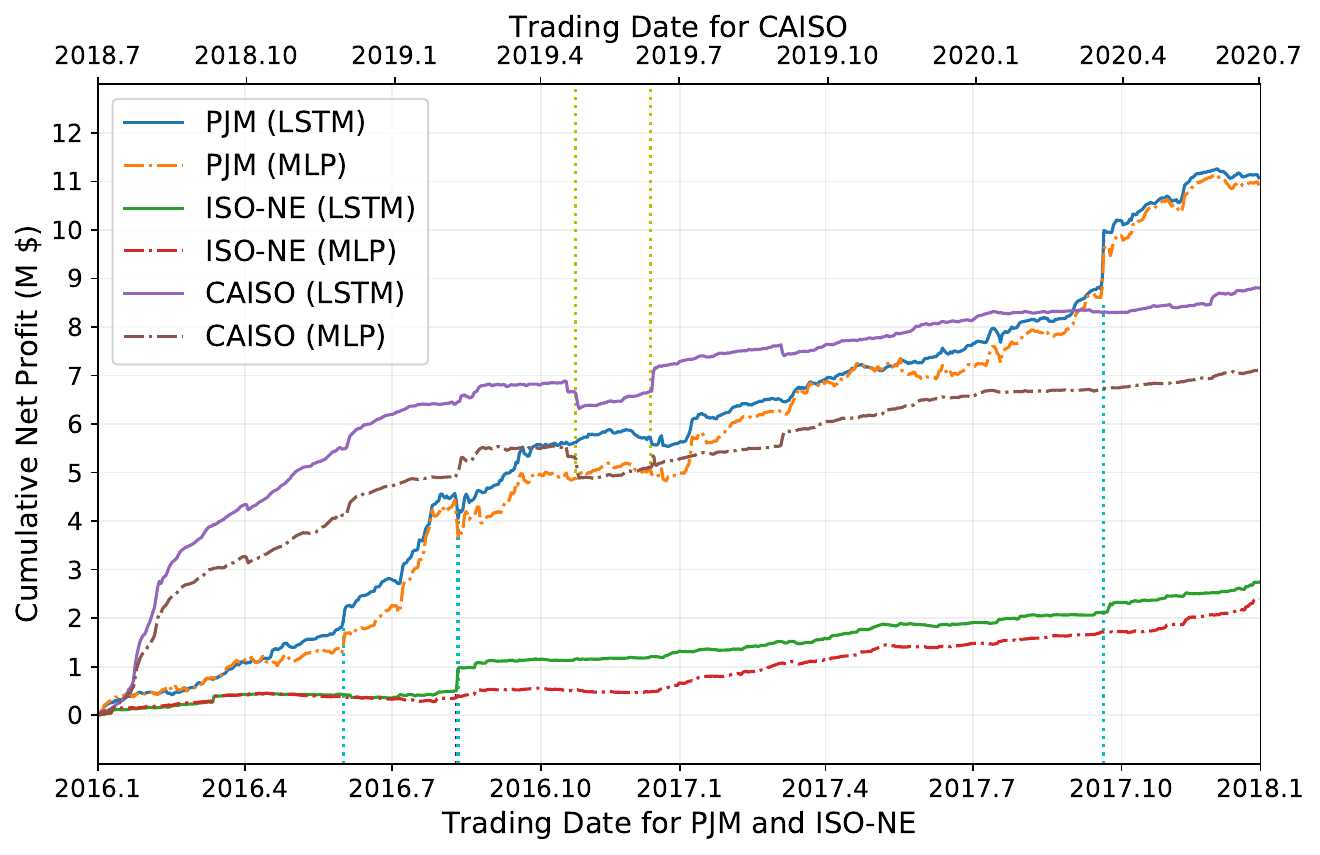}
    \caption{Cumulative net profits of algorithmic trading strategy without considering price sensitivity.}
    \label{fig:NoPriceSensitivity}
\end{figure}

The cumulative net profits of algorithmic trading strategy using MLP and LSTM models for the three electricity markets are depicted in Figure \ref{fig:NoPriceSensitivity}. As shown in the figure, the proposed algorithmic trading strategy is very profitable in all three electricity markets when the price sensitivity is not considered. When the LSTM model is used to predict LMP spreads, the algorithmic trading strategy yields approximately \$11M, \$9M, and \$3M of cumulative net profits for PJM, CAISO, and ISO-NE in a 2-year period.

Figure \ref{fig:NoPriceSensitivity} also shows that algorithmic trading strategy based on the LSTM model is much more profitable than that of the MLP model in CAISO market. In ISO-NE market, the LSTM algorithm slightly outperforms MLP. In PJM market, the cumulative net profit achieved by the LSTM-based and the MLP-based algorithmic trading strategies are roughly the same. In terms of net profit, the virtual bid portfolio derived from the LSTM-based LMP spread forecast outperforms MLP-based portfolio by 1.4\%, 14.1\% and 23.9\% for PJM, ISO-NE, and CAISO respectively. The share of non-hydro renewable generation in CAISO, 27\%, is much higher than that of PJM, 3\%. During early spring and summer days, CAISO is much more likely to experience significantly faster ramping in net-load than PJM. By considering the explanatory variables in the past few hours, the LSTM model is more capable of capturing the potential shortage in supply and LMP spikes than the MLP model in CAISO market.

Note that rare market events could lead to dramatic virtual bid portfolio gains and losses. In the PJM market, a significant gain of \$0.95M and a notable loss occurred on Sep. 20, 2017 and Aug. 12, 2016 respectively due to high temperature and peak load conditions. In ISO-NE, an unusual gain of \$0.65M took place on Aug. 11, 2016 due to severe generation capacity deficiency caused by thunder storms. In CAISO, notable portfolio gain and loss happened on Apr. 18 and Jun. 10, 2019 respectively. These two events are caused by the unforeseen and sharp drop in renewable generation.

\subsection{Impact of Portfolio Risk Limit on the Profitability of the Algorithmic Trading Strategy}

\begin{figure}[!t]
    \centering
    \includegraphics[width=3.5in]{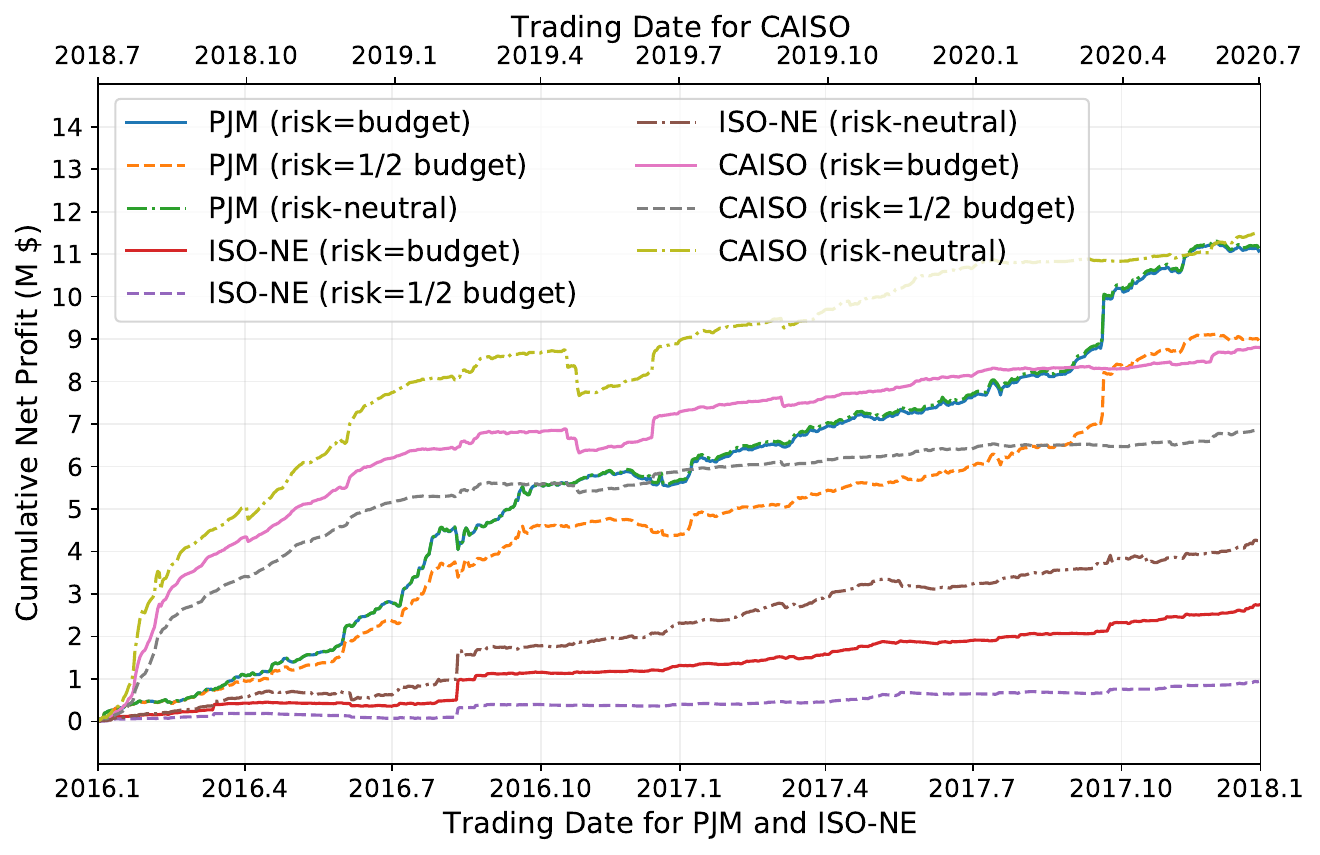}
    \caption{Cumulative net profits of virtual bid trading strategy under different risk limits without considering price sensitivity.}
    \label{fig:LSTM_risk_comparison}
\end{figure}

This subsection evaluates the impact of portfolio risk limit on the profitability of the algorithmic virtual bid trading strategy. We conduct the virtual bid portfolio optimization under three risk limit scenarios ranging from risk-averse to risk-neutral. In scenario 1, the portfolio risk level equals to one half of the portfolio budget limit. In scenario 2, the portfolio risk level equals to the portfolio budget limit. In scenario 3, we completely remove the risk limit, which makes the proprietary trading company risk neutral. The cumulative net profits of the virtual bid trading strategy under three risk limit scenarios are calculated based on LSTM model and depicted in Fig. \ref{fig:LSTM_risk_comparison}.

As shown in Fig. \ref{fig:LSTM_risk_comparison}, by focusing solely on potential gains regardless of the risk, the risk-neutral portfolio achieves notably higher cumulative net profits than both risk-averse portfolios for ISO-NE and CAISO. For PJM, when we increase the risk limit from one half of the portfolio budget to the portfolio budget, the cumulative net profit increases significantly. When we further relax the risk constraint by removing it, the improvement in net profit becomes negligible. This is because the budget constraint rather than the risk limit constraint is binding most of time for the portfolio optimization under scenario 2 in PJM. When we reduce the risk limit from the budget limit to one half of the budget limit, the cumulative net profits decrease by 18\%, 60\%, and 22\% respectively for PJM, ISO-NE, and CAISO.
 

\subsection{Profitability of Algorithmic Trading Strategies Considering Price Sensitivity}
In this subsection, we quantify the profitability of algorithmic trading strategy considering price sensitivity. Here we analyze two scenarios. In both scenarios, when reporting net profit, the impact of the trading company's virtual bids on the LMP is taken into consideration. The first scenario is called the full price sensitivity scenario, where the trading company considers the impacts of its virtual bids on LMP spread while solving the portfolio optimization problem. The second scenario is called the partial price sensitivity scenario, where the trading company does not consider price sensitivity while performing virtual bid portfolio optimization.

\begin{figure}[!t]
    \centering
    \includegraphics[width=3.5in]{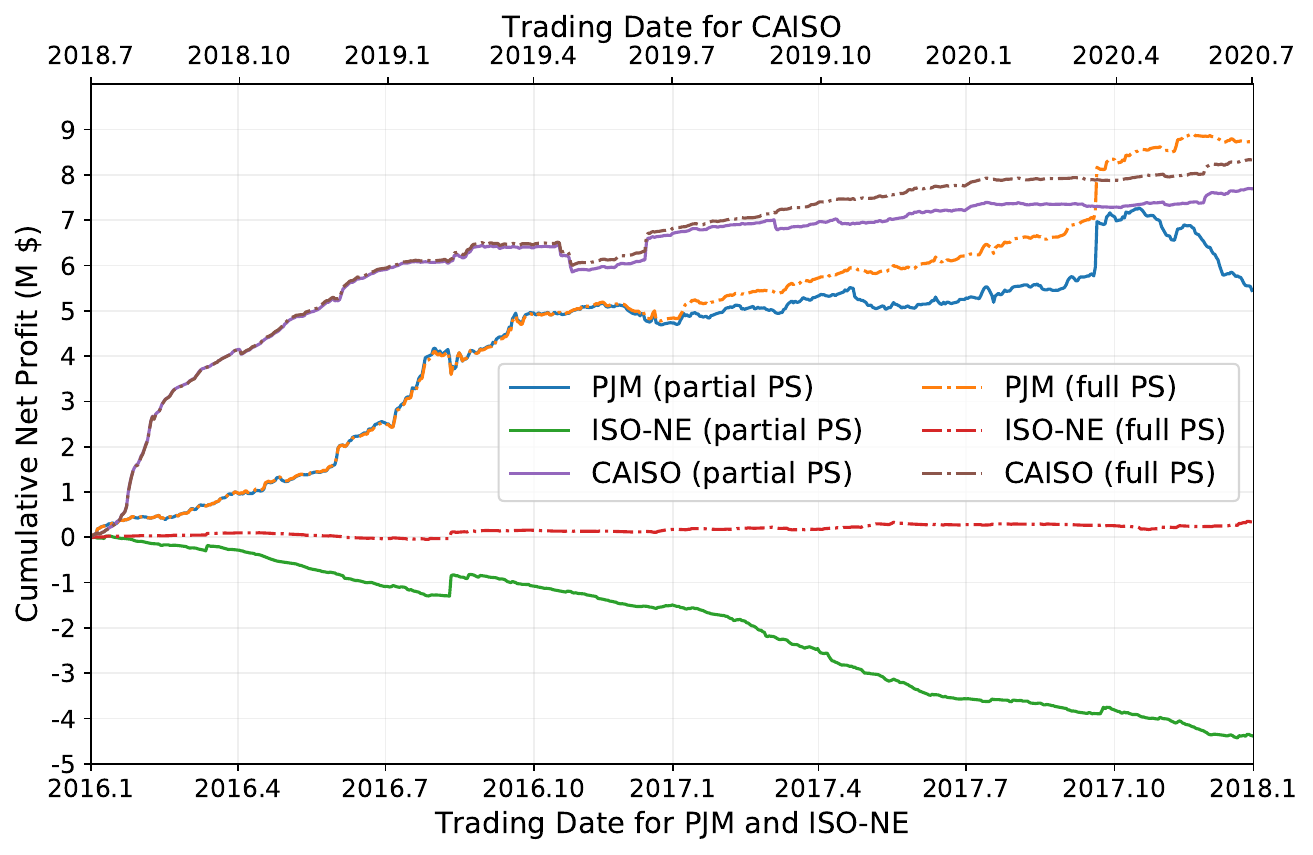}
    \caption{Cumulative net profits of algorithmic trading strategy considering the impact of virtual bids on LMP when reporting profit and loss. ``Full PS" refers to the full price sensitivity scenario and ``partial PS" refers to the partial price sensitivity scenario.}
    \label{fig:LSTM-PriceSensitivity}
\end{figure}
\begin{figure}[!t]
    \centering
    \includegraphics[width=3.5in]{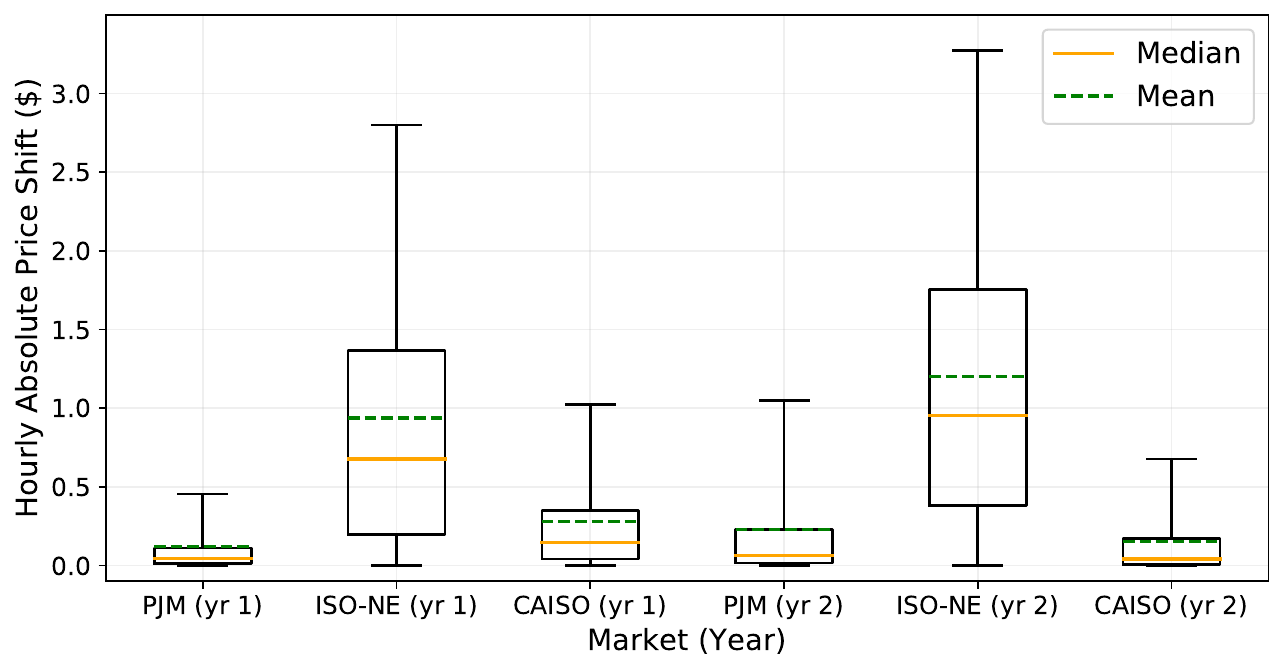}
    \caption{Statistics of hourly price shift in absolute value. This shows the hourly change of price spreads averaged on each year considering the price sensitivity. The percentiles here are set to 5\% and 95\%.}
    \label{fig:HourlyPriceShift}
\end{figure}

Figure \ref{fig:LSTM-PriceSensitivity} depicts the cumulative net profits of algorithmic trading strategy with LSTM model under the full and partial price sensitivity scenarios. By comparing the scenario without price sensitivity in Fig. 2 and the full price sensitivity scenario in Fig. 3, it can be seen that the trading company's virtual bidding activity reduce a sizable portion of its algorithmic trading strategies' cumulative net profits. The reductions in cumulative net profits in PJM, CAISO and ISO-NE are approximately \$2 Million, \$0.7 Million, and \$2.5 Million. By comparing the full and partial price sensitivity scenarios in Fig. 3, we conclude that the proposed virtual bids portfolio optimization strategy that considers price sensitivity explicitly performs much better than the one that ignores price sensitivity. The differences in cumulative net profit between the full and the partial price sensitivity scenarios is the largest for ISO-NE (\$4 Million), followed by PJM (\$2.5 Million), and CAISO (\$0.9 Million). This is because the impacts of virtual bids on LMP spread is the largest in ISO-NE due to its small market size and large price sensitivity. Figure \ref{fig:HourlyPriceShift} shows the box plot of the changes in LMP spread due to the trading company's virtual bids that correspond to 5\% of virtual bid market share verifies the statement above. As shown in Fig. \ref{fig:HourlyPriceShift}, the median LMP spread change in ISO-NE is much higher than that of PJM and CAISO.


{Table \ref{table:price_convergence} shows the convergence of LMP considering price sensitivity with 5\% virtual bid market share. As shown in the table, the presence of virtual bid does lead to reduction in LMP spread between DA and RT markets.
\begin{table}[!t]
    \setlength\extrarowheight{2pt}
    \renewcommand{\arraystretch}{1.2}
    \caption{LMP Convergence considering Price Sensitivity}
    \label{table:price_convergence}
    \centering
    \begin{tabular}{c|c|c|c}
    \hlinewd{1pt}
    \multirow{2}{*}{Market} & \multirow{2}{*}{Year} & \multicolumn{2}{c}{Average Absolute LMP Spread (\$/MWh)} \\
                            \cline{3-4}
                            &                       & without virtual bidding & with virtual bidding \\
    \hlinewd{1pt}
    \multirow{2}{*}{PJM} & yr 1 & 6.37 & 6.25 \\
                         \cline{2-4}
                         & yr 2 & 6.07 & 5.84 \\
    \hlinewd{1pt}
    \multirow{2}{*}{ISO-NE} & yr 1 & 9.39  & 8.45 \\
                         \cline{2-4}
                         & yr 2 & 10.38 & 9.18 \\
    \hlinewd{1pt}
    \multirow{2}{*}{CAISO} & yr 1 & 13.25 & 12.97 \\
                         \cline{2-4}
                         & yr 2 & 7.11 & 6.96 \\
    \hlinewd{1pt}
    \end{tabular}
\end{table}
}

\subsection{Efficiency Analysis of Two-settlement Power Markets}
In this subsection, we evaluate the efficiency of three wholesale power markets' two-settlement system by measuring the performance of virtual bid portfolio with different market shares. The first performance metric quantifies the cumulative net profit of virtual bids portfolio per dollar of collateral and risk limit. The second metric is the Sharpe ratio which is often used in finance to measure the performance of an investment portfolio. Specifically, Sharpe ratio measures the performance of an investment portfolio compared to a risk-free asset after adjusting for its risk. The Sharpe ratio $S_p$ of an investment portfolio can be calculated as:
\begin{gather*}
    S_p = \dfrac{E[R_p-R_f]}{\sqrt{var[R_p-R_f]}}
\end{gather*}
where $R_a$ is the portfolio's rate of return, $R_f$ is the rate of return of the risk-free asset. If the algorithmic virtual bidding strategy yields a portfolio with higher performance, the two-settlement system of the corresponding wholesale market is expected to have lower efficiency. 

To vary the market share of the virtual bidding portfolio, we choose different daily portfolio budgets. Note that the risk limit is set equal to the daily portfolio budget limit. From the perspective of the trading company, we assume that the virtual bid portfolio can take market shares from 1\% to 10\% with approximately 1\% stepsize.

\begin{figure}[!t]
    \centering
    \includegraphics[width=3.5in]{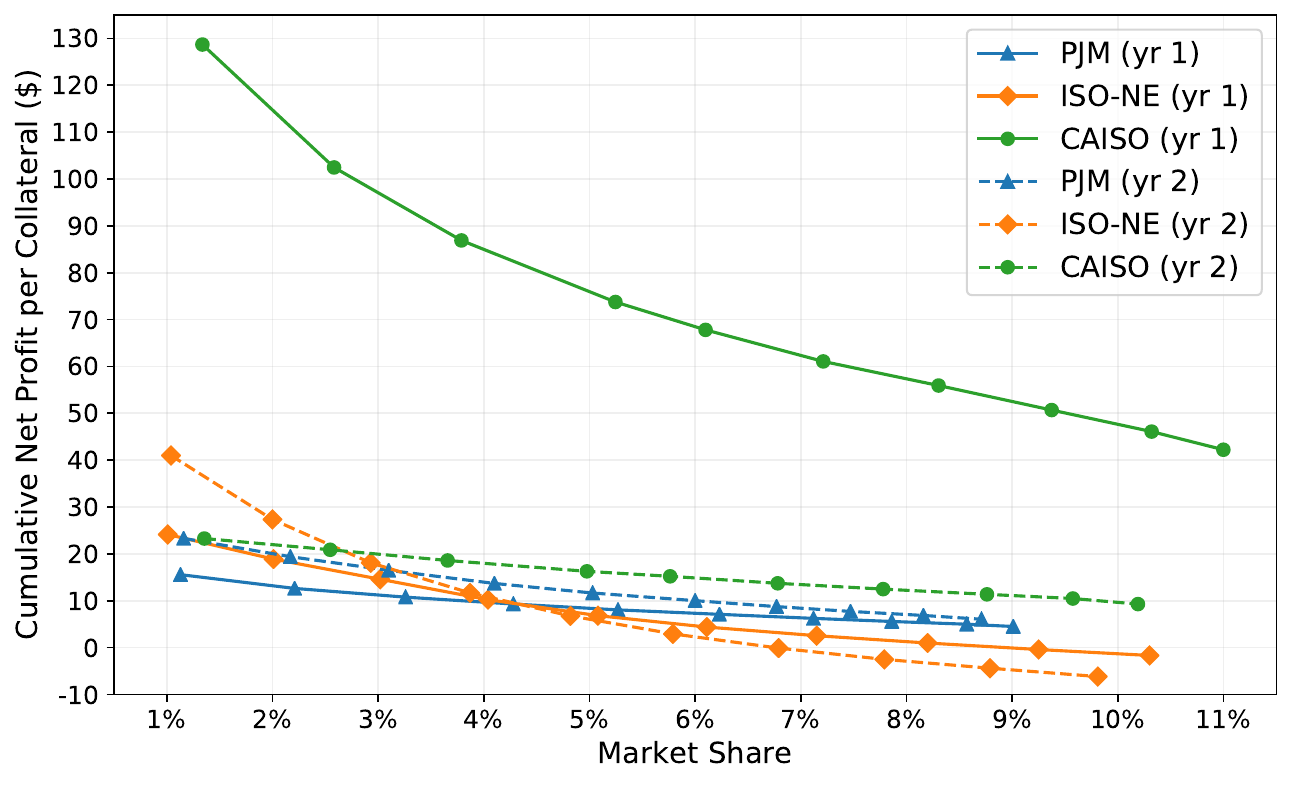}
    \caption{Profitability of virtual bids with different market shares.}
    \label{fig:MarketProfitability}
\end{figure}

Figure \ref{fig:MarketProfitability} shows the annual cumulative portfolio net profit per dollar of budget and collateral from the three markets in two separate years with ten different market shares. It can be observed that the net profit per dollar decreases as the market share increases. This result can be explained by two main reasons. 
First of all, after taken the most profitable bidding positions, only less profitable virtual bids can be identified with increased portfolio budget. Secondly, margins between DA and RT LMPs decrease as trading quantities increase, which further slows down the increase in portfolio net profit.

CAISO's two settlement system is shown to be the least efficient among the three wholesale markets, as the algorithmic trading strategy achieves the highest profitability in CAISO market during the second year of study. 
The LSTM model is able to forecast the LMP spreads in CAISO with 70\% accuracy, compared with 58\% and 60\% for PJM and ISO-NE (accuracy here refers to the ratio of correctly-predicted directions of LMP spreads, either positive or negative). With 1\% market share, the algorithmic trading strategy secured a profitability of up to \$41 and \$23 per dollar of collateral in ISO-NE and PJM. The reason of the profitability being the lowest in PJM is that it is the most competitive market with the largest number of virtual traders and trading volume.

To better understand the market efficiency and portfolio performance, we calculate the annual virtual bid portfolio's Sharpe ratio with different market shares. In contrast with the first portfolio performance metric, the Sharpe ratio measures the performance of the virtual bids portfolio compared to a risk-free asset after adjusting for its risk. If the algorithmic trading strategy achieves a high Sharpe ratio, then the corresponding market's efficiency should be low.

\begin{figure}[!t]
    \centering
    \includegraphics[width=3.5in]{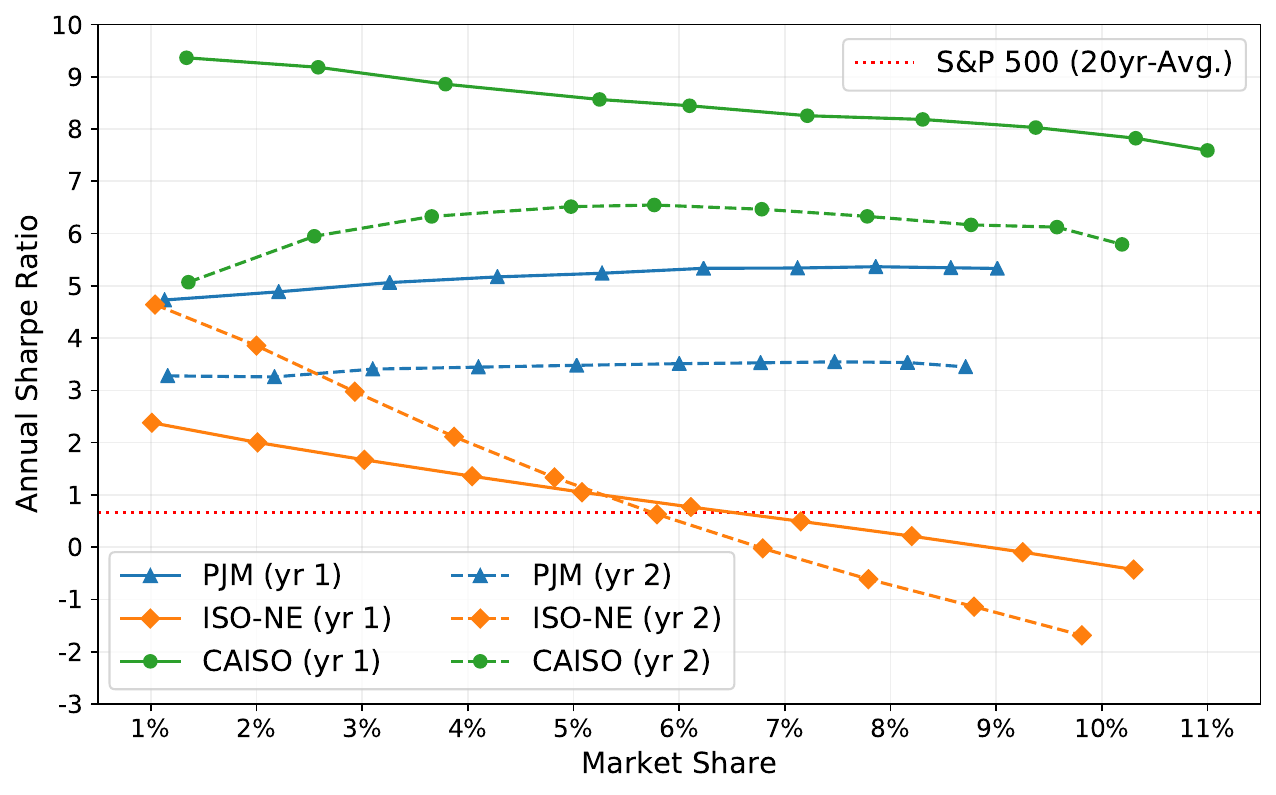}
    \caption{Sharpe ratios of algorithmic virtual bids trading portfolios.}
    \label{fig:SharpeRatio}
\end{figure}

Figure \ref{fig:SharpeRatio} illustrates the Sharpe ratios of virtual bid portfolio in three power markets and S\&P 500 index for the corresponding years. With a wealth of historical data, we report the 20-year average Sharpe ratio for the S\&P 500 index, which is a stock market index measuring the stock performance of 500 large companies listed on stock exchanges in the United States.
The Sharpe ratio of virtual bid portfolios for CAISO and PJM are much higher than that of the S\&P 500 index for all market shares. When the ISO-NE's virtual bid portfolio's market share is 5\% of below, its Sharpe ratio is also higher than that of S\&P 500 index. This indicates that the electricity markets' two settlement systems are in general much less efficient than the stock market. According to the Sharpe ratios, CAISO has the least efficient two settlement system among the three wholesale power markets. Note that the Sharpe ratio of PJM virtual bid portfolio is higher than that of ISO-NE. This result is different from the portfolio profitability curves shown in Figure \ref{fig:MarketProfitability}. This because our proposed algorithmic trading strategy captures the extremely high price spread between DA and RT LMPs on August 11, 2016, which results in a 1100\% daily return. 
It significantly increases the total net profit, but reduces the Sharpe ratio which penalizes volatility in portfolio returns.

\section{Conclusion}
This paper develops an algorithmic virtual bid trading strategy that considers the impacts of virtual bids on LMPs. A constrained gradient boosting tree is developed to model the monotonic function representing the sensitivity of LMP spread. The risk-constrained virtual bid portfolio optimization problem is reformulated as a mixed-integer quadratically-constrained problem via convex relaxation. The results of comprehensive empirical studies on the three U.S. electricity markets show that the proposed virtual bid portfolio optimization framework considering price sensitivity outperforms the one that ignores it. Among the three U.S. electricity markets, the proposed algorithmic virtual bid trading strategy achieves the highest profit in CAISO.  The Sharpe ratios of virtual bid portfolios for PJM, ISO-NE, and CAISO are all significantly higher than that of S\&P 500 index when the virtual bidder's market share is lower than 5\%. Given the high uplift cost of net virtual supply in U.S. wholesale electricity markets such as CAISO, it would be interesting to develop algorithmic trading strategies to exploit the differences in congestion patterns between day-ahead and real-time markets.
\ifCLASSOPTIONcaptionsoff
  \newpage
\fi



\bibliographystyle{IEEEtran}
\bibliography{./citation}
%



%
\begin{IEEEbiography}[{\includegraphics[width=1in,height=1.25in,clip,keepaspectratio]{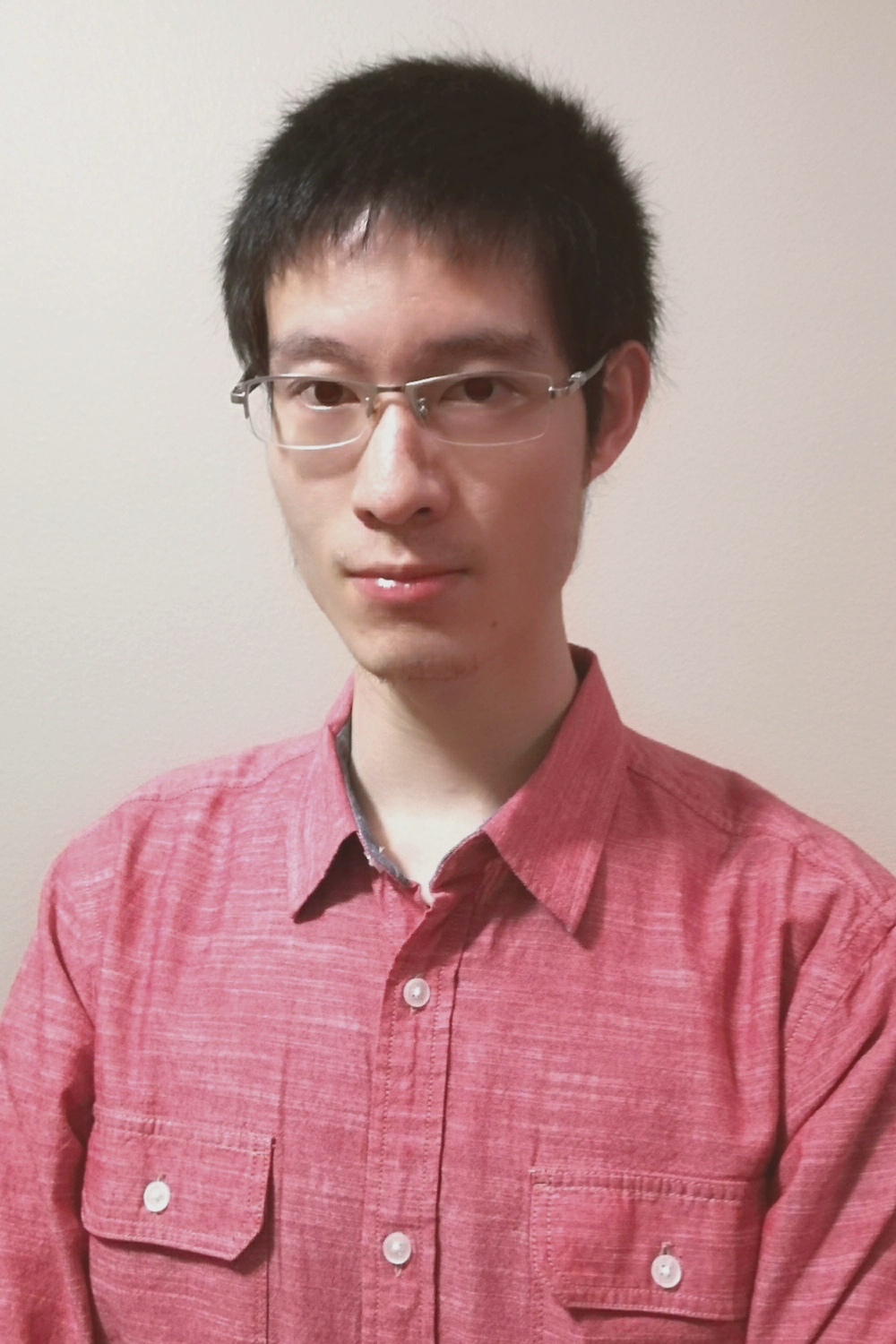}}]{Yinglun Li}
(S'20) received the B.S. degree in Information Engineering from the Xi'an Jiaotong University, Xi'an, China and the M.S. degree in Computer Engineering from the University of California, Riverside, USA in 2017 and 2019 respectively. He is currently pursuing the Ph.D. degree in Electrical and Computer Engineering at University of California, Riverside, USA. His research interests include data analytics and machine learning applications in electricity markets.
\end{IEEEbiography}

\begin{IEEEbiography}[{\includegraphics[width=1in,height=1.25in,clip,keepaspectratio]{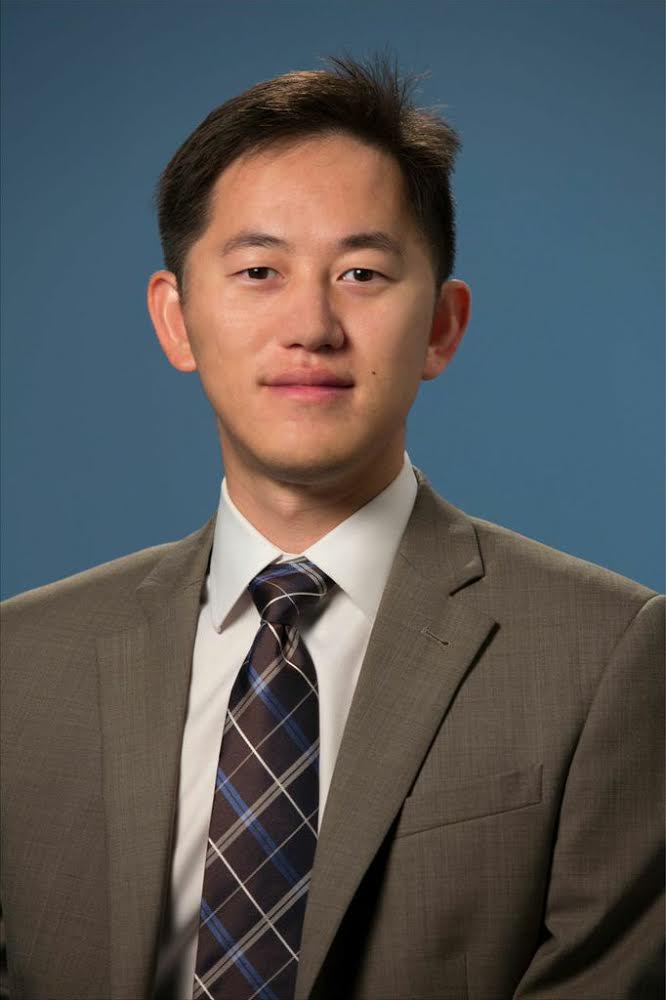}}]{Nanpeng Yu}
(M'11-SM'16) received his B.S. in Electrical Engineering from Tsinghua University, Beijing, China, in 2006. Dr. Yu also received his M.S. and Ph.D. degree in Electrical Engineering from Iowa State University, Ames, IA, USA in 2007 and 2010 respectively. He is currently an Associate Professor in the department of Electrical and Computer Engineering at University of California, Riverside, CA, USA. His current research interests include machine learning theory, big data analytics in smart grid, electricity market design, and smart energy communities. Dr. Yu is an Editor of IEEE Transactions on Smart Grid, IEEE Transactions on Sustainable Energy, and International Transactions on Electrical Energy Systems.
\end{IEEEbiography}

\begin{IEEEbiography}[{\includegraphics[width=1in,height=1.25in,clip,keepaspectratio]{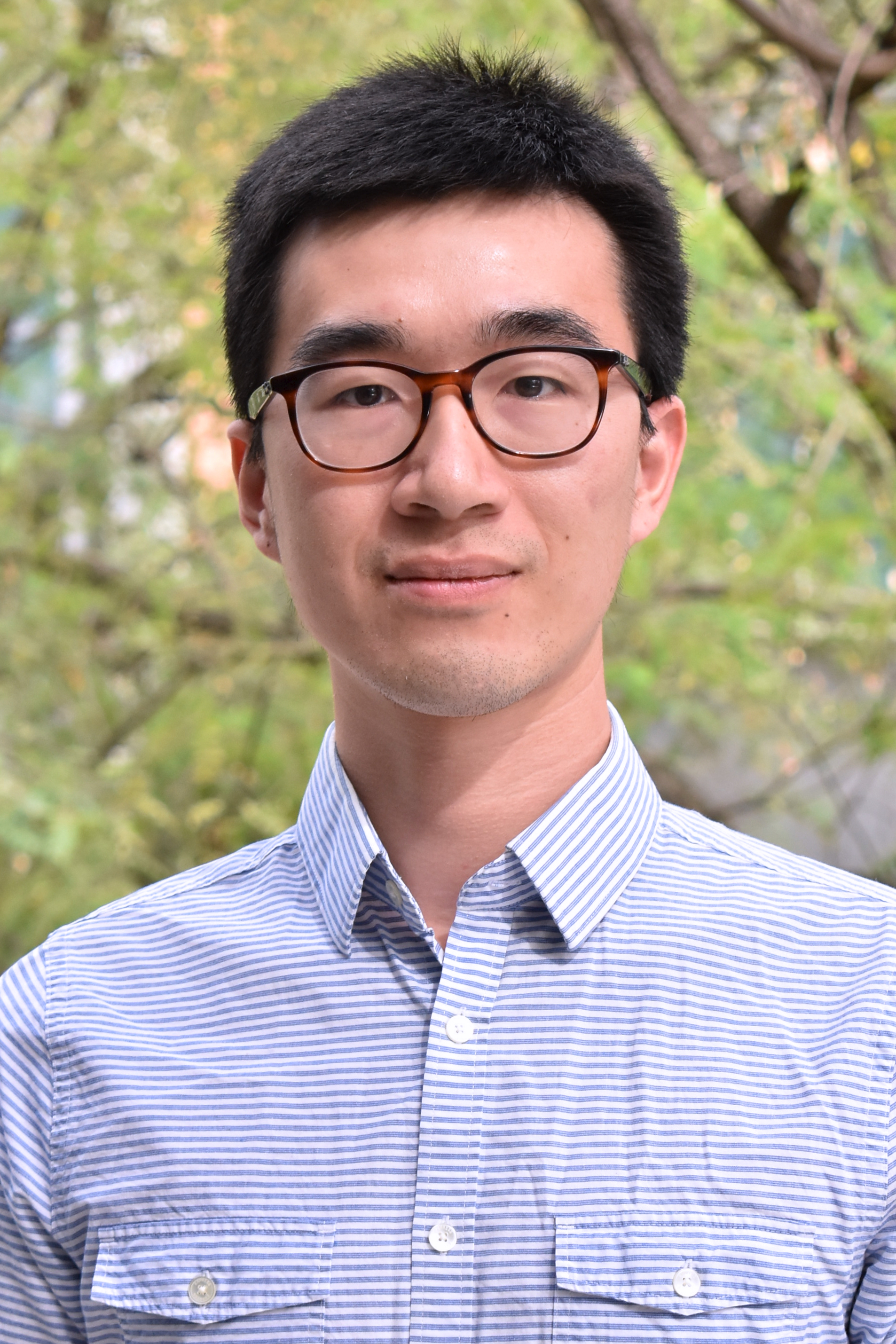}}]{Wei Wang}
(S'17) received the B.S. degree in Electrical Engineering from the Huazhong University of Science and Technology, Wuhan, China and the M.S.E. degree in Electrical and Computer Engineering from the University of Michigan, Ann Arbor, USA in 2012 and 2014 respectively. He received the Ph.D. degree in Electrical and Computer Engineering at University of California, Riverside, USA in 2020. His research interests include optimizations and machine learning applications in power systems. He is currently a senior data scientist in Walmart Labs.
\end{IEEEbiography}









\end{document}